# The distribution of information content in English sentences

Shuiyuan Yu[1], Jin Cong[2], Junying Liang[2], Haitao Liu[2*]

**Abstract**

*Sentence is a basic linguistic unit, however, little is known about how information content is distributed across different positions of a sentence. Based on authentic language data of English, the present study calculated the entropy and other entropy-related statistics for different sentence positions. The statistics indicate a three-step staircase-shaped distribution pattern, with entropy in the initial position lower than the medial positions (positions other than the initial and final), the medial positions lower than the final position and the medial positions showing no significant difference. The results suggest that: (1) the hypotheses of Constant Entropy Rate and Uniform Information Density do not hold for the sentence-medial positions; (2) the context of a word in a sentence should not be simply defined as all the words preceding it in the same sentence; and (3) the contextual information content in a sentence does not accumulate incrementally but follows a pattern of "the whole is greater than the sum of parts".*

Human language is a communication system for information transmission. Previous research has confirmed that language phenomena of various types are inseparably related to information transmission. For instance, a linguistic unit (e.g., word, syllable, phoneme or prosodic unit) tends to exhibit reduced articulation when its predictability increases (i.e., its information content or entropy decreases) due to the influence of its context (1-7). In language production, speakers tend to insert words with little or no contribution to utterance meaning (e.g., the relativizer *that*) or use the full forms of words and phrases (e.g., *information*, *you are*) in more information-dense parts of sentence to reduce information density; conversely, they are more likely to omit words with little or no meaning or use reduced forms of words and phrases (e.g., *info*, *you're*) in less information-dense parts of sentence to increase information density (8-12). Moreover, the information content of a word has been found to be directly proportional to its length and thus a better predictor of the latter than word frequency (13, 14). Cross-linguistic statistics have also revealed a negative correlation between information density and speech rate in communication (15). The number of bits conveyed per word is claimed to be a determinant of cognitive load measures including reading time (16). Of the six basic word orders of human language, the object-first languages are found to be the least optimized from the perspective of information density (17), which reasons why only less than 2% of the world's languages are of this type (18).

[1] School of Computer, Communication University of China, Beijing, 100024, P.R. China. [2] Department of Linguistics, Zhejiang University, Hangzhou, 310058, P.R. China.

[*] To whom correspondence should be addressed. E-mail: htliu@163.com

In an effort to explicate the above phenomena, theories have been proposed. Among them one is the Constant Entropy Rate (CER), claiming that speakers tend to keep the entropy rate of speech at a constant level (19); and the other is Uniform Information Density (UID) (8, 9), supposing that the language-generating system prefers an even distribution of the intended message in the speech stream. The major tenet of these two theories consists with the conclusion of information theory, that is, the most efficient way of transmitting information through a noisy channel is at a constant rate close to the capacity of the channel (20). If the communicative properties of human language have already been optimized through evolution, human language would have a stable rate of information transmission. In other words, the uniformity of information density may be a constraint on the production and comprehension of natural language.

As a basic unit of natural language, sentence has always been an important subject of research in various disciplines. However, little is known about the role of information content (entropy) in sentence processing. In other words, we still lack a definite answer to such a basic question as how information content is distributed within a sentence. In addition to syntactic and semantic factors, information content may constitute another important constraint on sentence production and comprehension. The intra-sentence distribution of information content (entropy) is not only relevant to various syntactic phenomena but also reflects the optimal efficiency of communication of the language system.

Currently there exist two major lines of research related to the intra-sentence distribution of information content. One line holds that CER and UID can be considered as general principles in sentence processing, and based on them we may anticipate that the information content in sentence is distributed uniformly. The other line focuses on the intelligibility and predictability of words in different positions of a sentence. The main findings of the latter line suggest that the sentence-final position may carry more information content than the other positions (hereafter as the "sentence-final effect"). For instance, Behaghel (21) found that longer words tend to occur later in a German sentence and the same is true of important, less predictable words. Rubenstein and Pickett (22, 23) claimed that the intelligibility and predictability of words in the sentence-final position tend to be different from other positions. The results of these two lines evidently diverge. More importantly, neither line of research has provided a systematic view of intra-sentence distribution of information content. The former line has only reached very general conclusions concerning information content of words in sentences but has not narrowed down the scope of research to the intra-sentence distribution of information content. Moreover, this line usually adopts the n-gram model for the quantification of information content of a word in sentence. However, this model only considers the n-1 words preceding the target word instead of the target word's position in the sentence. The latter line, although shedding light on the intelligibility and predictability of words in the sentence-final positions, lacks systematic investigation of the information content of different positions in a sentence and only relies on indirect observations which are not amenable to quantification. In addition, as the observations of this line of research are not based on large collections of authentic natural language data, the conclusions are open to test.

The current goal is to obtain a systematic and quantitative characterization of the intra-sentence distribution of information content based on a large body of authentic language data of modern English. We calculated a series of entropy-related statistics for different positions in English sentences, including the positional entropy (1-gram entropy), number of words (types), power-law exponent of the words' probability distribution, mean word length, mean word frequency, and proportions of different word-frequency classes. As sentences vary in length, the same position in different sentences may play different roles in their intra-sentence distributions of information content. Therefore, we calculated the statistics for sentences with a length (in the number of words) from 3 to 50, inclusive, respectively.

**Results**

We first calculated the 1-gram entropy (noted as *H(X)*) for different sentence positions. Basically, it is an out-of-context entropy, for it is dependent exclusively upon sentence positions without considering the specific context. This entropy signifies the mean information content of all the words occurring in a given sentence position, hence it is the conditional entropy given sentence positions in strict terms. Here, we refer to this entropy as *positional entropy*.

Figure 1 shows the positional entropy (*H(X)*) in different positions of sentences with a length of 15, 30 and 45. It is noteworthy that *H(X)* follows a three-step staircase-shaped distribution across different sentence positions, with the sentence-initial position significantly lower than the medial positions (positions other than the initial and final), the medial positions significantly lower than the final position, and the medial positions showing no significant difference. For sentences with a length of 15, the correlation coefficient between *H(X)* and sentence-medial position is 0.4770, the F-statistic of *H(X)* in different medial positions 2.3569 and its probability 0.1633. For sentences with a length of 30 and 45, the corresponding

results are -0.0607, 0.0851, 0.7731 and -0.0606, 0.1399, 0.7104, respectively (see supporting information for the results of other classes of sentence length).

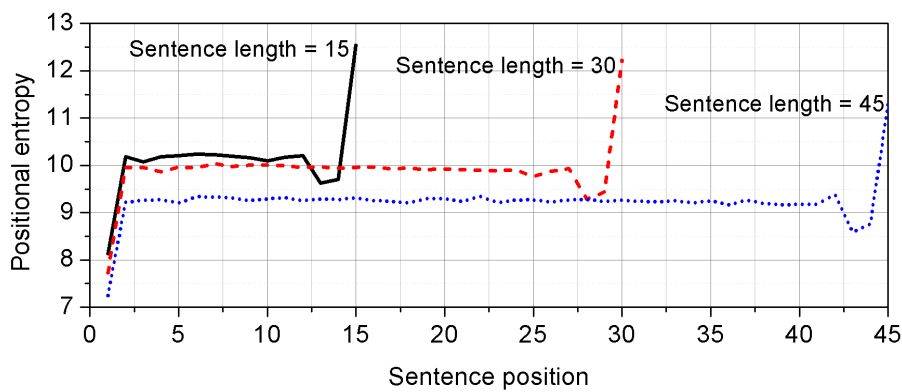

Figure 1 Positional entropy in different sentence positions

According to Formula 1 (see 'Materials and Methods'), the positional entropy *H(X)* of a given sentence position is dependent upon the number of words (types) occurring in this position. The greater the number of words (types) in this position, the greater the value of *H(X)* will be. Also, it is dependent upon the probability distribution of these words. The more uniform the distribution, the greater the value of *H(X)*. Given these two considerations, we examined the number of words (types) and the probability distribution of these words for different sentence positions. As the probability distribution of words in any given sentence position generally follows power-law distribution and the power-law exponent is always negative, this power-law exponent was adopted here as a measure for the uniformity of the probability distribution.

Similar to Figure 1, the number of words (types) and the power-law exponent of the words' probability distribution also exhibit a three-step staircase-shaped distribution across different sentence positions. Further, no significant change was found in the sentence-medial positions (see supporting information for the results in detail).

Of studies concerning information content of human language, the most influential are perhaps the investigations of the correlation between information content and word length (24, 25). It has been found that information content exhibits a strong positive correlation with word length (13, 14), that is, longer words tend to transmit more information. Hence, it is anticipated that calculating the change of mean word length with sentence position can reflect the distribution of information content in a sentence.

Quite in line with our prediction, the relationship between mean word length and sentence position exhibits the same pattern as illustrated in Figure 1 (see supporting information for the results in detail).

Moreover, the mean word frequency for different sentence positions was also adopted as a statistic for the intra-sentence distribution of information content. As high-frequency words tend to be more predictable than the low-frequency ones and the latter constitute majority of the lexicon (see 'Materials and Methods' for information concerning the words' rank-frequency distribution in the language data of the present study), mean word frequency can be adopted as another indicator of information content in a given sentence position. Figure 2 displays the mean word frequency in different positions of sentences with a length of 15, 30 and 45 (see supporting information for results of other classes of sentence length). The mean word frequency in the sentence-initial position is higher than the medial positions, the medial positions higher than the final position, and the final position significantly lower than any other position. As a higher mean word frequency means less information content, Figure 2 again reveals a three-step staircase-shaped distribution of intra-sentence information content. The results in Figure 2 are consistent with Otto Behaghel's (21) finding that important, less predictable words tend to occur later in a sentence.

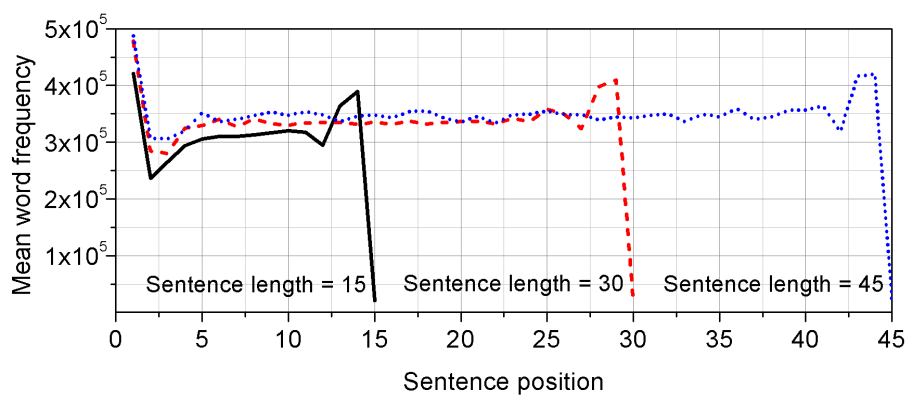

Figure 2 Mean word frequency in different sentence positions.

Besides, the intra-sentence distribution of information content can also be observed by means of word-class-related statistics. As high-frequency sequences and syntactic relations are more likely to be grammaticalzed (26-31), high-frequency words tend to be function words. "[T]he roles of words and other linguistic phenomena such as morphology, phonology, and syntax are highly influenced by low, medium, or high frequency with which they occur" (26). Given this, certain word classes can be roughly distinguished from the others by their differences in word frequency. In the present study, we classified the words into three classes of frequency, namely high (the first 100 in the word frequency rank), low (the last 100 in the rank) and medium (the rest). High-frequency words are mostly function words, which carry less information content; while medium- and low-frequency words are generally content words, with more information content. The intelligibility of words at a given sentence position can be quantified as the ratio of contents words to function words (22). The proportions of words (tokens) of different frequency classes in different sentence positions were calculated as an indicator of the intra-sentence distribution of information content. As displayed in Figure 3 are the proportions of different word-frequency classes in different positions of sentences with a length of 15 (left), 30 (middle) and 45 (right) (see supporting information for the results of other classes of sentence length).

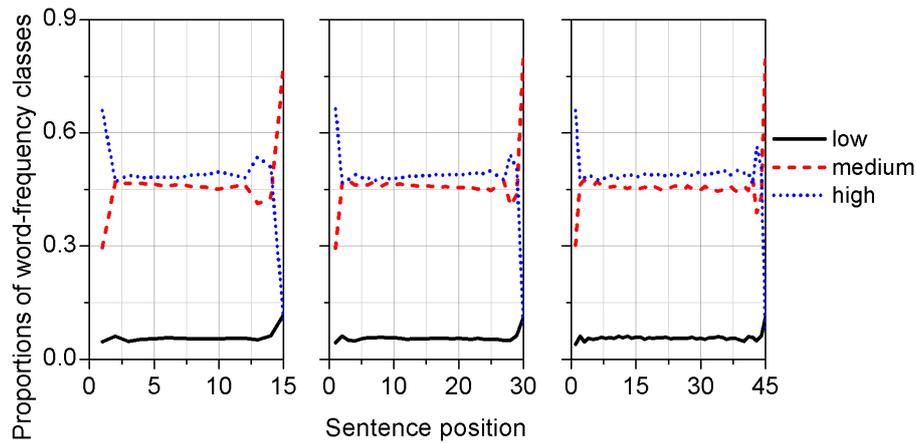

Figure 3 The proportions of words (tokens) of three frequency classes in different sentence positions.

The sentence-initial position exhibits the greatest proportions of high-frequency words. The proportions of all three word-frequency classes are rather fixed in the medial positions, with that of high-frequency words (proportion ≈48%) greater than the medium-frequency (proportion ≈42%) and low-frequency (proportion ≈10%) ones. The medium- and low-frequency words form the majority in the sentence-final position, with their proportions significantly higher than high-frequency words. The behavior of the sentence-final position in this respect is consistent with what Rubenstein and Pickett (22) have found. Again, these proportions in different sentence positions indicate a three-step staircase-shaped distribution of information content in the sentences.

As the frequency of a word is negatively correlated with its information content, the change of its frequency with respect to sentence position can also reflect the intra-sentence distribution of information content. The current statistics further reveal that for the sentence-medial positions, the frequency of some words (proportion = 50.15%) increases with sentence position while that of others (proportion = 49.85%) exhibits the reversed tendency (see supporting information for the results in detail). The proportions of these two types of words exhibit no significant difference, as confirmed by the F-statistic 0.36 and the corresponding probability 0.5497. This indicates that the out-of-context information content does not fluctuate significantly in the sentence-medial positions.

**Discussion**

The present study investigated how information content is distributed across different positions of a sentence. To realize, we calculated the entropy and other entropy-related statistics for different sentence positions. Our results explicitly demonstrate a three-step staircase-shaped distribution with respect to sentence position. Specifically, these statistics increase substantially only in the sentence-initial and sentence-final positions but do not exhibit significant change with respect to the sentence-medial positions.

It is found that the words occurring in the sentence-initial position are predominantly high-frequency words. The number of words (types), mean word length and positional entropy in this position are all smaller than the medial positions. The words in the medial positions are predominantly medium- and high-frequency words, and the number of words (types), mean word length and positional entropy in these positions do not exhibit a significant difference but are all larger than the initial position. The words in the sentence-final position are predominantly medium- and low-frequency words and the proportion of low-frequency words manifest a considerable increase compared with the preceding positions. The number of words (types), mean word length and positional entropy all reach a peak in this position.

In addition, the stable level of entropy and other entropy-related statistics in the sentence-medial positions does not necessarily mean that they increase without statistical significance. Indeed, many of these statistics tend to decrease with respect to the sentence-medial positions (see supporting information for the results in detail).

As suggested by the information theory, the relationship between positional entropy $H(X)$, contextual information content $I(X, C)$ and conditional entropy $H(X/C)$ are given by $H(X) = I(X, C) + H(X/C)$ (See Formula 2 and its detailed explanations in 'Materials and Methods').

Given the above relationship, the results of the current study can be interpreted as follows: the sum of contextual information content and conditional entropy exhibits a three-step staircase-shaped increase with respect to sentence position, with a substantial increase in the sentence-initial and sentence-final positions and the medial positions showing no significant difference.

The notion of entropy is crucial to linguistic studies, for it is capable of measuring the difficulty of information processing. Theoretically, to achieve the optimal efficiency of communication, the entropy during language processing should be maintained at a constant level as postulated by the hypothesis of CER. As for the contextual information content *I(X, C)*, up to now there have been no solid findings concerning how it is distributed within a sentence. A straightforward approach is that the contextual information available to a word in a sentence is mainly from the words preceding it. That is to say, the contextual information content in a sentence is supposed to increase with sentence position.

Our results, based on authentic language data and in light of Formula 2, help illuminate the relationship between contextual information content *I(X, C)* and conditional entropy *H(X/C)*.

On the one hand, suppose that the entropy *H(X/C)* is constant in language processing as postulated in CER, our findings would mean that the contextual information content *I(X, C)* increases with sentence position with the same pattern as *H(X)*. That is, *I(X, C)* in the sentence-initial position is less than the medial positions, the medial positions less than the final position, and the medial positions exhibit no significant difference. The stable level of *I(X, C)* in the medial positions means that a word does not get more contextual information than the words in the preceding positions. In other words, the context of a word in a sentence should not be simply defined as all the words preceding it in the same sentence.

In addition, if we assume that CER is valid, the uniform distribution of contextual information content in the sentence-medial positions and the notable increase of contextual information content in the final position would indicate that the contextual information content does not accumulate for the greater part of a sentence and follows a pattern of "the whole is greater than the sum of parts" during the course of the sentence, no matter how context is defined.

On the other hand, if the context of a word in a sentence is defined as all the words preceding it, then the more the preceding words, the greater $I(X, C)$ is. Given the increase of $I(X, C)$ and the relationship in Formula 2, it can be inferred that the entropy $H(X/C)$ would decrease with the sentence-medial positions. However, our results have shown that $H(X)$ maintains at a stable level in the sentence-medial positions, That is to say, CER does not hold for the sentence-medial positions, such that UID does not work in this case.

To note, the three-step staircase-shaped increase of positional entropy suggests that the conditional probability of a word occurring in a specific position in a sentence does not have "position" homogeneity.

With regard to the conflict, we are more inclined to assume the validity of CER (also UID) rather than model the context of a word in a sentence as all the words preceding it. For one thing, both CER and UID have a solid theoretical basis upon information theory (9, 19). For another, assuming the validity of CER (also UID), our results will be consistent with a large number of research findings (21-23) and our intuition about language.

By assuming the validity of CER and UID, our results can be interpreted as follows. In the sentence-initial position, the sentence has only started to unfold and the contextual information available to this position is at its minimum. This explains why the initial position exhibits the least positional entropy, the smallest number of words (types) and the highest proportion of high-frequency words. It has been found that high-frequency words tend to be hubs of a linguistic system from a perspective of complex network (32, 33) and are more easily accessed. In the sentence-final position, the whole sentence has virtually been completed and the words in the preceding positions provide contextual information and sufficient sentence structure information, such that words with greater out-of-context information content are used here to keep the in-context information content of words in this position consistent with the preceding words. As content words carry greater information content, they are predominant in the sentence-final position. Meanwhile, as the words preceding the sentence-final position are mostly medium- and high-frequency words, which usually do not have strong relevance to the sentence topic, words with stronger relevance to the sentence topic need to be filled in the final position to make the meaning of the whole sentence complete. Generally, the candidates for the final position are low-frequency, content words. As the range of sentence topics is infinite, the number of content words (types) that can occur in the final position is huge. This echoes what we have found in the current study, and is also consistent with the sentence-final effect (21) and the findings concerning the intelligibility and predictability of sentence-final words (22, 23).

Our current findings about the relationship between contextual information and sentence positions can potentially contribute to modeling the context of words in sentences and push forward the understanding of the Markov model of context as widely-used at present.

**Materials and Methods**

The language data we used in the present study is part of the British National Corpus (BNC) (34), with 17,293 texts. These texts contain altogether 328,935 word types, 34,853,585 word tokens with 3,372 classes of word frequency (i.e., with the highest word frequency rank 3,372). The total number of sentences is 1,956,195 (all occurrences of sentence counted) and 1,857,414 (identical sentences counted only once). The result of power-law fitting to the rank-frequency distribution of the words is $s(p) = 1.581e+05*p^{-1.835}+11.78$, whereby p is word frequency, s the number of words (types) with the given frequency. The 95% confidence bounds of the three

parameters in the fitting equation are (1.579e+05, 1.582e+05), (-1.839, -1.832) and (8.962, 14.59), respectively. The determination coefficient of the power-law fitting (adjusted R$^2$) is 0.9992.

The current study adopts the positional entropy and other entropy-related statistics, instead of calculating information content directly. The reason for this is that the accurate calculation of information content is far from easy (35) due to two features. On the one hand, it is difficult to quantify a word's contextual information (especially mean-related information), which is indeed essential to the calculation of information content. On the other hand, the information content estimates of sequences with long range correlations (including natural languages) converge very slowly with the sequence length (36, 37).

As suggested by Ferrer-i-Cancho et al. (38), positional entropy is closely related with UID and CER. If UID holds for words in sentences and we assume that contextual information increases with sentence position, then the information content or entropy out of context should increase with sentence position. Therefore, we can examine whether information content has an even intra-sentence distribution based on the change of entropy out of context with respect to sentence position.

The positional entropy (1-gram) of a given sentence position is given by

$$H(X) = -\sum p_i \log_2 p_i, \qquad (1)$$

whereby $H(X)$ is the 1-gram entropy of the position, the random variable $X$ any possible word (type) that can occur in this position, $p_i$ the probability of $X$=word $i$. This Formula is adapted from Formula (1) of Shannon's work (36) and $H(X)$ of the former is equivalent to F1 of the latter. As seen from Formula 1, $H(X)$ is an out-of-context entropy, for it is dependent exclusively upon sentence position instead of the specific context.

According to information theory, the relationship between positional entropy, contextual information content and conditional entropy is given by

$$H(X) = I(X, C) + H(X/C), \qquad (2)$$

whereby *H(X)* represents the positional entropy of a given position, the random variable *X* any word (type) occurring in the position, the random variable *C* the context of *X*, *H(X/C)* the conditional entropy given context *C* (i.e., the entropy as in CER), and *I(X, C)* the mutual information between context *C* and the position.

The statistics we calculated generally exhibit a three-step staircase-shaped distribution in the sentences and do not fluctuate significantly in the sentence-medial positions, which is confirmed by their correlations with sentence position and the F-statistics plus the corresponding probabilities (see supporting information for the results in detail). Considering the divergent behaviors of the second, third, penultimate and antepenultimate positions as indicated by almost all of these statistics, the calculation of the correlation coefficients and F-statistics plus the corresponding probabilities for the sentence-medial positions excluded these four positions.

**Acknowledgments**

This work is partly supported by the National Social Science Foundation of China (Grant No. 11&ZD188).

## References

1. van Son R, Koopmans-van Beinum FJ, Pols LCW (1998) Efficiency as an organizing principle of natural speech. *ICSLP98*, p 0.
2. van Son R, van Santen J (2005) Duration and spectral balance of intervocalic consonants: A case for efficient communication. *Speech Communication* 47: 100–123.
3. Bolinger D (1963) Length, vowel, juncture. *Linguistics* 1: 5–29.
4. Lieberman P (1963) Some effects of semantic and grammatical context on the production and perception of speech. *Lang Speech* 6: 172–187.
5. Aylett MP (1999) Stochastic suprasegmentals: Relationships between redundancy, prosodic structure and syllabic duration. *Proceedings of the 1999 International Conference on Spoken Language Processing* pp. 289-292.
6. Pluymaekers M, Ernestusb M, Baayen RH (2005) Lexical frequency and acoustic reduction in spoken Dutch. *J Acoust Soc Am* 118: 2561–2569.
7. Bell A, Brenier J, Gregory M, Girand C, Jurafsky D. (2009) Predictability effects on durations of content and function words in conversational English. *Journal of Memory and Language* 60: 92–111.
8. Jaeger TF (2006) Redundancy and syntactic reduction in spontaneous speech. PhD thesis

(Stanford University, Palo Alto, CA).
9. Levy R, Jaeger TF (2007) Speakers optimize information density through syntactic reduction. *Advances in Neural Information Processing Systems 19*, eds Schölkopf B, Platt J, Hoffman T (MIT Press, Cambridge, MA), pp 849–856.
10. Aylett M, Turk A (2004) The smooth signal redundancy hypothesis: A functional explanation for relationships between redundancy, prosodic prominence, and duration in spontaneous speech. *Lang Speech* 47:31–56.
11. Mahowald K, Fedorenko E, Piantadosi ST, Gibson E. (2013) Info/information theory: speakers choose shorter words in predictive contexts. *Cognition* 126:313-8.
12. Frank AF, Jaeger TF (2008) Speaking rationally: Uniform information density as an optimal strategy for language production. *The 31st Annual Meeting of the Cognitive Science Society*, eds Love BC, McRae K, Sloutsky VM (Cognitive Science Society, Austin, TX), pp 939–944.
13. Manin D (2006) Experiments on predictability of word in context and information rate in natural language. *Journal of Information Processes* 6:229–236.
14. Piantadosi ST, Tily H, Gibson E (2011) Word lengths are optimized for efficient communication. Proc Natl Acad Sci USA 108:3526–3529.
15. Pellegrino F, Coupé C, Marsico E (2011) Across-language perspective on speech information rate. *Language* 87: 539–558.
16. Hale J (2003) The information conveyed by words in sentences. *J Psycholinguist Res* 32:101–123.
17. Maurlts L, Perfors A, Navarro D (2010) Why are some word orders more common than others? A uniform information density account. *Advances in Neural Information Process Systems* 23: 1585-1593.
18. Tomlin RS (1986) *Basic word order: functional principles* (Croom Helm, London).
19. Genzel D, Charniak E (2002) Entropy rate constancy in text. *Proceedings of the 40th annual meeting of the association for Computational linguistics* pp. 199–206.
20. Shannon C (1948) A mathematical theory of communications. *Bell Systems Technical Journal* 27: 623 – 656.
21. Behaghel O (1909) Beziehungen zwischen Umfang und Reihenfolge von Satzgliedern. *Indogermanische Forschungen* 25: 110-142.
22. Rubenstein H, Pickett JM (1958) Intelligibility of Words in Sentences. *J Acoust Soc Am* 30: 670-670.
23. Rubenstein H, Pickett JM (1957) Word Intelligibility and Position in Sentence. *J Acoust Soc Am* 29: 1263-1263.
24. Ball P (2011) How words get the message across. Nature News, 10.1038/news.2011.40. Available at http://www.nature.com/news/2011/110124/full/news.2011.40.html.
25. Griffiths TL (2011). Rethinking language: How probabilities shape the words we use. *Proc Natl Acad Sci USA* 108: 3825-3826.
26. Bybee J (2007) *Frequency of Use and the Organization of Language* (Oxford University Press, New York).
27. Bybee J (2003) Mechanisms of change in grammaticization: The role of frequency. *The Handbook of Historical Linguistics* eds Joseph BD, Janda RD (Blackwell, Oxford), pp 602–623.
28. Hopper PJ, Traugott EC (1993) *Grammaticalization* (Cambridge University Press,

Cambridge).
29. Bybee J, Hopper P (2001) Introduction to frequency and the emergence of linguistic structure. *Frequency and the Emergence of Linguistic Structure* eds Bybee J, Hopper P (John Benjamins, Amsterdam), pp 1–24.
30. Haspelmath M (2008) Creating economical morphosyntactic patterns in language change. *Language universals and language change* ed Good J (Oxford University Press, Oxford), pp 185–214.
31. Haspelmath M (2008) Frequency vs. iconicity in explaining grammatical asymmetries. *Cogn Linguist* 19: 1–33.
32. Liu H (2008) The complexity of Chinese syntactic dependency networks. *Physica A* 387:3048--3058.
33. Solé RV, Corominas-Murtra B, Valverde S, Steels L (2010) Language networks: Their structure, function and evolution. *Complexity* 15: 20-26.
34. Clear J (1993) *The British National Corpus* (MIT Press, Cambridge, MA).
35. Paninski L (2003) Estimation of entropy and mutual information. *Neural Comput* 15: 1191–1253.
36. Shannon C (1951) Prediction and entropy of printed English. *The Bell System Technical Journal* 30: 50-64.
37. Schurmann T, Grassberger P (1996) Entropy estimation of symbol sequences. *Chaos* 6: 414–427.
38. Ferrer-i-Cancho R, Dębowski Ł, Moscoso del Prado Martín F (2013) Constant conditional entropy and related hypotheses. Journal of Statistical Mechanics L07001.

(2013-07-19, the first version)